%% file: main.tex
\documentclass[letterpaper, 10 pt, conference]{ieeeconf}  % Comment this line out
\IEEEoverridecommandlockouts                              % This command is only
\usepackage{amsmath} % assumes amsmath package installed
\usepackage{amssymb}  % assumes amsmath package installed
\usepackage{mathtools}
\usepackage{cite}

\usepackage{nicefrac}

\usepackage{tikz}
\usepackage{pgfplots}
\usepackage{tikzscale}
\usepackage{siunitx}
\usepackage{multirow}
\usepackage[font=small]{caption}
\usepackage[font=small]{subcaption}
\usepackage{graphicx}

\usepackage{hyperref}
\usepackage[capitalise]{cleveref}

\let\labelindent\relax
\usepackage{enumitem}

\usepackage{amssymb}% http://ctan.org/pkg/amssymb
\usepackage{pifont}% http://ctan.org/pkg/pifont
\newcommand{\cmark}{\ding{51}}%
\newcommand{\xmark}{\ding{55}}%
\usepackage{booktabs}
\newtheorem{cor}{Corollary}

\usepackage{mathtools}

\DeclarePairedDelimiter\norm{\lVert}{\rVert}%

\usepackage{packages/herofigure}
\newif\ifarXiv
\arXivtrue % show arXiv version
\title{\LARGE \bf
Optimization-Based System Identification and Moving Horizon Estimation Using Low-Cost Sensors for a Miniature Car-Like Robot
}

\author{Sabrina Bodmer$^\star$, Lukas Vogel$^\star$, Simon Muntwiler, Alexander Hansson,\\ Tobias Bodewig, Jonas Wahlen, Melanie N. Zeilinger, and Andrea Carron  % <-this % stops a space
\thanks{$^\star$Sabrina Bodmer and Lukas Vogel contributed equally to this work.}
\thanks{The work of Simon Muntwiler was supported by the Bosch Research Foundation im Stifterverband. The authors are members of the Institute for Dynamic Systems and
Control, ETH Zurich, Switzerland.}%
}

\begin{document}

\maketitle

\thispagestyle{empty}
\pagestyle{empty}

%%%%%%%%%%%%%%%%%%%%%%%%%%%%%%%%%%%%%%%%%%%%%%%%%%%%%%%%%%%%%%%%%%%%%%%%%%%%%%%%
\begin{abstract}
This paper presents an open-source miniature car-like robot with low-cost sensing and a pipeline for optimization-based system identification, state estimation, and control.
The overall robotics platform comes at a cost of less than \$\,700 and thus significantly simplifies the verification of advanced algorithms in a realistic setting.
We present a modified bicycle model with Pacejka tire forces to model the dynamics of the considered all-wheel drive vehicle and to prevent singularities of the model at low velocities.
Furthermore, we provide an optimization-based system identification approach and a moving horizon estimation (MHE) scheme.
In extensive hardware experiments, we show that the presented system identification approach results in a model with high prediction accuracy, while the MHE results in accurate state estimates.
Finally, the overall closed-loop system is shown to perform well even in the presence of sensor failure for limited time intervals.
All hardware, firmware, and control and estimation software is released under a BSD 2-clause license to promote widespread adoption and collaboration within the community.
\end{abstract}

%%%%%%%%%%%%%%%%%%%%%%%%%%%%%%%%%%%%%%%%%%%%%%%%%%%%%%%%%%%%%%%%%%%%%%%%%%%%%%%%

\input{sections/introduction}

%%%%%%%%%%%%%%%%%%%%%%%%%%%%%%%%%%%%%%%%%%%%%%%%%%%%%%%%%%%%%%%%%%%%%%%%%%%%%%%%

\input{sections/hardware}

%%%%%%%%%%%%%%%%%%%%%%%%%%%%%%%%%%%%%%%%%%%%%%%%%%%%%%%%%%%%%%%%%%%%%%%%%%%%%%%%

\section{SYSTEM AND SENSORS MODELLING}\label{sec:modelling}
In this section, we describe the system and sensor models of the Chronos car used for system identification and state estimation. Furthermore, we introduce the model discretization.\looseness=-1
\subsection{System Model}
We consider a dynamic bicycle model with a simplified Pacejka tire force model~\cite{rajamani2011vehicle, Carron2023}, with adaptations similar to~\cite{Raji2022} to account for AWD configurations. The overall system state is denoted as $x=\begin{bmatrix}  x_{\mathrm{p}} & y_{\mathrm{p}} & \psi & v_{\mathrm{x}} & v_{\mathrm{y}} & \omega \end{bmatrix}^\top \in \mathbb{R}^{n_{\mathrm{x}}}$ with $n_{\mathrm{x}}=6$. The~$x$ and~$y$ coordinates, as well as the yaw angle in world frame, are given by $x_{\mathrm{p}}, y_{\mathrm{p}}$ and $\psi$. Longitudinal and lateral velocities, as well as the yaw rate, are denoted by $v_{\mathrm{x}}, v_{\mathrm{y}}$ and $\omega$, and are given in body frame. The model input $u=\begin{bmatrix} \delta & T \end{bmatrix}^\top \in \mathbb{R}^{n_{\mathrm{u}}}$ with $n_{\mathrm{u}}=2$ consists of the steering angle $\delta$ and the input torque $T$.
The continuous-time system is governed by the following differential equations
\begin{subequations}\label{eq:dynamic}
\allowdisplaybreaks
	\begin{align}
		\dot{x}_{\mathrm{p}} &= v_{\mathrm{x}} \cos(\psi) - v_{\mathrm{y}} \sin(\psi),\\
		\dot{y}_{\mathrm{p}} &= v_{\mathrm{x}} \sin(\psi) + v_{\mathrm{y}} \cos(\psi),\\
		\dot{\psi} &= \omega,\\
		\dot{v}_{\mathrm{x}} &= \frac{1}{m} \left( F_{\mathrm{x,r}} + F_{\mathrm{x,f}}\cos(\delta) - F_{\mathrm{y,f}} \sin(\delta) \right. \\
  & \qquad \left. + mv_{\mathrm{y}} \omega - F_{\mathrm{fr}}\right), \nonumber \\
		\dot{v}_{\mathrm{y}} &= \frac{1}{m} \left( F_{\mathrm{y,r}} + F_{\mathrm{x,f}}\sin(\delta) + F_{\mathrm{y,f}} \cos(\delta) - mv_{\mathrm{x}} \omega\right), \\
		\dot{\omega} &= \frac{1}{I_\mathrm{z}} \left( F_{\mathrm{y,f}} l_{\mathrm{f}} \cos(\delta) + F_{\mathrm{x,f}}l_{\mathrm{f}}\sin(\delta) - F_{\mathrm{y,r}} l_{\mathrm{r}}\right),
	\end{align}
\end{subequations}
where $m$ is the mass of the car, $I_\mathrm{z}$ is the inertia along the $z$-axis, and $l_{\mathrm{f}}$ and $l_{\mathrm{r}}$ are the distance of the front and rear axis from the center of mass, respectively.
\begin{figure}[t]
\centering
\resizebox{.9\linewidth}{!}{\input{figures/tikz/dynamic}}
\caption{Dynamic bicycle model with velocity in direction of each wheel used for the wheel encoder model (adapted from~\cite{Froehlich2022}).}  \vspace{-15px}
\label{fig:dynamic} 
\end{figure}
The lateral tire forces $F_{\mathrm{y,f}}$ and $F_{\mathrm{y,r}}$ are modeled with the simplified Pacejka tire force model
\begin{subequations}
	\begin{align*}
		F_{\mathrm{y,f}} &= D_{\mathrm{f}} \sin(C_{\mathrm{f}} \arctan(B_{\mathrm{f}} \alpha_{\mathrm{f}})), \\
		F_{\mathrm{y,r}} &= D_{\mathrm{r}} \sin(C_{\mathrm{r}} \arctan(B_{\mathrm{r}} \alpha_{\mathrm{r}})),
	\end{align*}
\end{subequations}
where $B_{\mathrm{f}}$, $B_{\mathrm{r}}$, $C_{\mathrm{f}}$,  $C_{\mathrm{r}}$, $D_{\mathrm{f}}$, and $D_{\mathrm{r}}$ are the Pacejka tire model parameters and  $\alpha_\mathrm{f}, \alpha_\mathrm{r}$ refer to the front and rear slip angles which depend on $v_{\mathrm{x}}$, $v_{\mathrm{y}}$, and $\omega$.
In the classical Pacejka model~\cite{Carron2023,rajamani2011vehicle}, the slip angles $\alpha_{\mathrm{f}}, \alpha_{\mathrm{r}}$ are discontinuous as $v_{\mathrm{x}} \rightarrow 0$, which poses a significant limitation when modeling the dynamics at low speeds or standstill.
To overcome this, we fit polynomials of third-degree ($p_{\mathrm{f}}, p_{\mathrm{r}}$) to express the slip angles in a low-velocity regime $|v_{\mathrm{x}}| \leq \epsilon$, with small constant $\epsilon \in \mathbb R_{>0}$. 
Therefore, the front and rear slip angles are computed as
\begin{subequations} \label{eq:dynamic_bicycle_lateral_force}
	\begin{align*}
		\alpha_{\mathrm{f}} &= \begin{cases}
                                    \hat\alpha_{\mathrm{f}}(v_{\mathrm{x}})= \arctan\left( \frac{-\omega l_{\mathrm{f}} - v_{\mathrm{y}}}{v_{\mathrm{x}}} \right) + \delta,& \text{if } |v_{\mathrm{x}}|\geq \epsilon,\\
                                    p_{\mathrm{f}}(v_{\mathrm{x}}) =b_{\mathrm{f}} v_{\mathrm{x}} + c_{\mathrm{f}} v_{\mathrm{x}}^3,              & \text{otherwise},
                                    \end{cases} \\
            \alpha_{\mathrm{r}} &= \begin{cases}
                                    \hat\alpha_{\mathrm{r}}(v_{\mathrm{x}}) = \arctan\left( \frac{\omega l_{\mathrm{r}} - v_{\mathrm{y}}}{v_{\mathrm{x}}} \right),& \text{if } |v_{\mathrm{x}}|\geq \epsilon,\\
                                    p_{\mathrm{r}}(v_{\mathrm{x}}) = b_{\mathrm{r}} v_{\mathrm{x}} + c_{\mathrm{r}} v_{\mathrm{x}}^3,              & \text{otherwise},
                                    \end{cases}
	\end{align*}
\end{subequations}
where $b_{\mathrm{f}}, c_{\mathrm{f}}, b_{\mathrm{r}}$, and $c_{\mathrm{r}}$ depend on $v_{\mathrm{y}}, \omega, l_{\mathrm{f}} $, and $l_{\mathrm{r}}$ and are determined such that the slip angle dynamics $\alpha_\mathrm{f}, \alpha_\mathrm{r}$ are continuously differentiable by solving 
\begin{subequations}
\begin{align*}
    \hat\alpha_{i}(\epsilon) &= p_{i}(\epsilon), \nonumber \quad 
    \frac{\partial}{\partial v_{\mathrm{x}}}\left(\hat\alpha_{i}(v_{\mathrm{x}}) \right)\bigg |_{v_{\mathrm{x}}=\epsilon} &= b_{i} + 3c_{i}\epsilon^2, \nonumber
\end{align*}
\end{subequations}
for $i=\{\mathrm{f},\mathrm{r}\}$.
We propose a modification of the standard model~\cite{rajamani2011vehicle, Carron2023}, as similarly done in~\cite{Raji2022}, where the longitudinal forces acting in the direction of the rear and front wheels, $F_{\mathrm{x,r}}$ and $F_{\mathrm{x,f}}$, respectively, are modeled as
%\begin{subequations}
%    \begin{align*}
$
        F_{\mathrm{m}} = (C_{\mathrm{m,1}}-C_{\mathrm{m,2}} v_{\mathrm{x}}) T,\ %\\
        F_{\mathrm{x,r}} = \gamma F_{\mathrm{m}},\ %\\
        F_{\mathrm{x,f}} = (1 - \gamma) F_{\mathrm{m}},
$
%\end{align*}
%\end{subequations}
where $C_{\mathrm{m, 1}}$, and $C_{\mathrm{m, 2}}$ are model parameters, and the factor~$\gamma$ is used to split the force between rear and front axle. Values of $\gamma = 1$, $\gamma=0$ and $0<\gamma<1$ can model rear-, front-, and all-wheel drive vehicles, respectively. We model friction effects in a parameterized Taylor approximation, inspired by the physical effects of roll resistance and viscous drag as
%\begin{equation*}
%    \label{eq:friction}
$
    F_{\mathrm{fr}} = \operatorname{sgn}(v_{\mathrm{x}}) (C_{\mathrm{d,2}} v_{\mathrm{x}}^2 + C_{\mathrm{d,1}} v_{\mathrm{x}} + C_{\mathrm{d,0}}),
$
%\end{equation*}
where $C_{\mathrm{d},i}$ are the expansion parameters.

The Pacejka tire model, physical model, friction, and motor parameters are given as 
$\theta_{\mathrm{p}} = [ D_{\mathrm{f}} \ D_{\mathrm{r}} \ C_{\mathrm{f}} \ C_{\mathrm{r}} \ B_{\mathrm{f}} \  B_{\mathrm{r}} ]$,  \\
 $\theta_{\mathrm{car}} = [ m  \ I_{\mathrm{z}} \ l_{\mathrm{f}} \ l_{\mathrm{r}}]$,
$\theta_{\mathrm{fr}} = [ C_{\mathrm{d,0}} \ C_{\mathrm{d,1}} \ C_{\mathrm{d,2}} ]$, 
and $\theta_{\mathrm{T}} = [ C_{\mathrm{m,1}} \ C_{\mathrm{m,2}}]$ respectively. The total dynamic model parameters are denoted as $\theta_{\mathrm{m}} = [\theta_{\mathrm{p}} \ \theta_{\mathrm{car}} \ \theta_{\mathrm{fr}} \ \theta_{\mathrm{T}}]$, with $n_{\theta_m}=15$.
\subsection{Sensor Models}
In the following, we will describe the sensor models of the inertial measurement unit (IMU), wheel encoders, and Lighthouse positioning system.

\subsubsection{Inertial measurement unit}
The IMU provides measurements of linear acceleration and angular velocity as
\begin{align*}
    y_{\mathrm{IMU}}=\begin{bmatrix}
        \frac{1}{m} \left( F_{\mathrm{x,r}} + F_{\mathrm{x,f}}\cos(\delta) - F_{\mathrm{y,f}} \sin(\delta) - F_{\mathrm{fr}}\right) + v_{\mathrm{y}} \omega \\
        \frac{1}{m}\left(F_{\mathrm{y,r}} + F_{\mathrm{x,f}}\sin(\delta) + F_{\mathrm{y,f}} \cos(\delta) \right) - v_{\mathrm{x}} \omega\\
        \omega
    \end{bmatrix}.
\end{align*}
The number of measurements is given as $ n_{\mathrm{IMU}} = 3$.

\subsubsection{Wheel encoders}
Custom-built wheel encoders are used to measure the angular velocity of each of the four wheels of the car. 
Assuming no skid, the measurement model of the wheel encoders can be obtained as
\begin{align}
    \label{eq:wheel-encoder-model}
    &y_{\mathrm{we}} = \frac{1}{r}\begin{bmatrix} v_{\mathrm{w,fl}} & v_{\mathrm{w,fr}} & v_{\mathrm{w,rl}} & v_{\mathrm{w,rr}} \end{bmatrix} ^\top = \\
    &\quad \frac{1}{r}\begin{bmatrix}
            \cos (\delta)(v_{\mathrm{x}} - 0.5 b_{\mathrm{car}} \omega) + \sin (\delta) (v_{\mathrm{y}} +  l_{\mathrm{f}}\omega) \\
            \cos (\delta)(v_{\mathrm{x}} + 0.5 b_{\mathrm{car}} \omega) + \sin (\delta) (v_{\mathrm{y}} +  l_{\mathrm{f}}\omega) \\
            v_{\mathrm{x}} - 0.5 b_{\mathrm{car}} \omega \\
            v_{\mathrm{x}} + 0.5 b_{\mathrm{car}} \omega \\        
    \end{bmatrix}, \nonumber
\end{align}
where $r$ is the wheel radius, $b_{\mathrm{car}}$ is the width of the car, and $v_{\mathrm{w,fl}}$, $v_{\mathrm{w,fr}}$, $v_{\mathrm{w,rl}}$, $v_{\mathrm{w,rr}}$, are the velocities in the direction of the front left, front right, rear left, and rear right wheels, respectively.
The parameters of the wheel encoders are denoted as $\theta_{\mathrm{we}} = \begin{bmatrix} r & b_{\mathrm{car}} \end{bmatrix}^\top \in \mathbb{R}^{n_{\theta_\mathrm{we}}}$, with $n_{\theta_\mathrm{we}} = 2$. The number of measurements is given as $ n_{\mathrm{we}} = 4$.

\subsubsection{Lighthouse positioning system}

The Lighthouse positioning system is used to measure the light plane impact angles $\alpha_{k,l}$, where $k \in \{1, 2, 3, 4\}$ refers to the sensor and $l \in \{1, 2\}$ determines the light plane. The position and orientation of the vehicle can be determined from these measurements. In the following, we provide the sensor model and relevant equations for a single base station. However, these can be extended to account for multiple base stations.
The measurement model can be obtained as
\begin{align*}
    &y_{\mathrm{lh}} = \begin{bmatrix} 
    \alpha_{1,1} &\alpha_{2,1} & \alpha_{3,1} & \alpha_{4,1} & \alpha_{1,2} & \alpha_{2,2} & \alpha_{3,2} & \alpha_{4,2}
    \end{bmatrix}, \nonumber
\end{align*}
where $\alpha_{k,l}$ is the impact angle for sensor $k$ and light plane~$l$.
The impact angles are given as
\begin{align}
    \label{eq:lighthouse-angle-measurement-model}
    \alpha_{k,l} = \tan^{-1}\left(\frac{y_{\mathrm{lh},k}}{x_{\mathrm{lh},k}}\right) + \sin^{-1}\left(\frac{z_{\mathrm{lh},k} \tan(t_{l})}{ \sqrt{x_{\mathrm{lh},k}^2 + y_{\mathrm{lh},k}^2}}\right), 
\end{align}
where $t_{1} = - \nicefrac{\pi}{6} - \delta t_{1}$ and $t_{2} = \nicefrac{\pi}{6} - \delta t_{2}$ are the tilt angles of the two light planes of each base station, and $\delta t_{1}$ and $\delta t_{2}$ are factory calibrated offsets. Finally, the position of each sensor ${}_{\mathcal L}p_{\mathrm{lh},k} = \begin{bmatrix} x_{\mathrm{lh},k} & y_{\mathrm{lh},k} & z_{\mathrm{lh},k} \end{bmatrix}^\top$, in the Lighthouse base station frame, is related to the position of the car via the transformation
%The position of each sensor $k$ in the frame of the base station is obtained as
\begin{align}
    {}_{\mathcal L}p_{\mathrm{lh},k} = R_{\mathrm{bs}}^\top \left(\begin{bmatrix} \cos (\psi) & -\sin (\psi) \\ \sin (\psi) & \cos (\psi) \\ 0 & 0 \end{bmatrix}\prescript{}{\mathcal{B}}p_{\mathrm{lh,k}}+ \begin{bmatrix} x_{\mathrm{p}} \\ y_{\mathrm{p}} \\ 0 \end{bmatrix} - p_{\mathrm{bs}} \right), \nonumber
\end{align}
where $\prescript{}{\mathcal{B}}p_{\mathrm{lh,k}}$ is the position of the $k$-th sensor in body frame and $R_{\mathrm{bs}}, p_{\mathrm{bs}}$ are the rotation and position of the Lighthouse base station. Thereby, $R_{\mathrm{bs}}$ is parametrized by the three rotation angles $\Phi = (\varphi_1, \varphi_2, \varphi_3)^\top$ as $R_{\mathrm{bs}} = R_z(\varphi_3) R_y (\varphi_2) R_x (\varphi_1)$.
The parameters of the Lighthouse model are denoted as $\theta_{\mathrm{lh}} = \begin{bmatrix} \Phi & p_{\mathrm{bs}} & \prescript{}{\mathcal{B}}p_{\mathrm{lh,k}} & \delta t_{1} & \delta t_{2} \end{bmatrix}^\top \in \mathbb{R}^{n_{\theta_\mathrm{lh}}}$, with $n_{\theta_\mathrm{lh}} = 16\cdot n_{\mathrm{bs}}$ and $n_{\mathrm{bs}}$ is the number of base stations. 
The number of measurements is given as $n_\mathrm{lh} = 8$. \looseness=-1

\subsection{Discrete-time System Model}

By numerically integrating~\eqref{eq:dynamic} and combining all sensor models, we obtain a discrete-time system of the form
\begin{subequations}\label{eq:dynamics_dt}
    \begin{align}
        x_{t+1} &= f(x_t,u_t,w_t,\theta), \label{eq:dynamics_dt_1} \\
        y_t & = h(x_t, u_t,w_t,\theta), \label{eq:dynamics_dt_2}
\end{align}
\end{subequations}
where $y \coloneqq \begin{bmatrix}  y_{\mathrm{IMU}}^\top & y_{\mathrm{we}}^\top & y_{\mathrm{lh}}^\top  \end{bmatrix}^\top \in \mathbb{Y} \subseteq \mathbb{R}^{n_{\mathrm{y}}}$ with $ n_{\mathrm{y}}\coloneqq n_{\mathrm{IMU}} + n_{\mathrm{we}} + n_{\mathrm{lh}}$ and $\theta \coloneqq \begin{bmatrix} \theta_{\mathrm{m}}^\top & \theta_{\mathrm{we}}^\top & \theta_{\mathrm{lh}}^\top  \end{bmatrix}^\top \in \mathbb{R}^{n_{\mathrm{\theta}}}$ with $n_{\theta} \coloneqq n_{\mathrm{\theta_{\mathrm{m}}}} + n_{\mathrm{\theta_{\mathrm{we}}}} + n_{\mathrm{\theta_{\mathrm{lh}}}}$. Additional process and measurement noise is denoted by $w_t \in \mathbb{W}\subseteq \mathbb{R}^{n_{\mathrm{w}}}$.
Note that $w_t$ appears in the dynamics~\eqref{eq:dynamics_dt_1} and measurement model~\eqref{eq:dynamics_dt_2} and hence can also model separate process disturbances and measurement noise.
We assume to know an estimate $\bar{x}_0$ of the initial state of the system~\eqref{eq:dynamics_dt} at time step $t=0$.
While the parameter $\theta$ is unknown, we assume to have access to a prior estimate $\Bar{\theta}_0$ and additional information in the form of constraints \looseness=-1
\begin{align} \label{eq:theta_constr}
    \theta \in \Theta \subseteq \mathbb{R}^{n_{\theta}}.
\end{align}
Note that if some parameters are easy to measure (e.g., the mass $m$), this can be enforced in the constraints~\eqref{eq:theta_constr}.
Additional information and physical state limits in the form of constraints $\mathbb X_{\mathrm{ID}}$ can be used during system identification, while it is often crucial to consider safety constraints on states~$\mathbb X$ and inputs~$\mathbb U$ during online operation.

%%%%%%%%%%%%%%%%%%%%%%%%%%%%%%%%%%%%%%%%%%%%%%%%%%%%%%%%%%%%%%%%%%%%%%%%%%%%%%%%
\section{SYSTEM IDENTIFICATION}
\label{sec:sys_ID}

To use the system model~\eqref{eq:dynamics_dt}, the model parameters~$\theta$ need to be identified.
In the following, we describe the calibration procedure for the Lighthouse positioning system and an optimization-based approach to identify the parameters $\theta$.

\subsection{Lighthouse Calibration}\label{sec:lh_calibration}

For practical reasons, we decouple determining the parameters $\theta_\mathrm{lh}$ from the system model identification. The parameters $\delta t_1, \delta t_2$ are factory-calibrated offsets of the light planes, provided in the base station's data stream. The position of the $k$ sensors in body frame, $_{\mathcal B}p_{\mathrm{lh}, k}$, can readily be measured.
To determine the position and rotation of the base station,~$p_{\text{bs}}$ and~$R_{\text{bs}}$, with respect to the world reference frame, we measure the Lighthouse angles received at~$n_\text{cal}$ static known positions~$p_i$, $i = 1, \ldots, n_\text{cal}$, and average them as $\bar \alpha_{i,l} = (\sum_{k} \alpha_{i,k,l})/k$, for both light planes $l \in \{ 1, 2 \}$.
We then solve an optimization problem minimizing the mean squared error between the angle measurements $\bar \alpha_i = \begin{bmatrix} \bar \alpha_{i, 1} & \bar \alpha_{i, 2} \end{bmatrix}^\top$ and the expected measurements from~\eqref{eq:lighthouse-angle-measurement-model},
\begin{equation}\label{eq:lighthouse-calibration-cost}
\min_{(\Phi, p_{\mathrm{bs}})} \sum_{i = 1}^{n_\text{cal}} \norm{\bar \alpha_i - \alpha_i(\Phi, p_{\mathrm{bs}}, p_i)}^2.
\end{equation}
The obtained Lighthouse parameters are denoted as $\hat{\theta}_{\mathrm{lh}}$.

\subsection{Optimization-based System Identification}
\label{sec:optimization-based-system-identification}

To estimate the parameters $\theta$ of the system~\eqref{eq:dynamics_dt} from noisy output measurements, we rely on an optimization-based approach similar to~\cite{bock1983recent, valluru2017development, simpson2023efficient}.
The (unknown) state trajectory of the system is denoted by~$\{ \hat x_j \}_{j = 0}^{L - 1}$. Assuming a set of input-output data of length $L$, $\mathbb{D} = \{ u_j, y_j\}_{0 = 1}^{L - 1}$, and an initial parameter estimate $\bar{\theta}_0$ are available, the objective is chosen as
\begin{align}
    V_{\mathrm{SysID}}(\hat{\theta}, \hat x_0, \{ \hat{w}_j\}_{j=0}^{L - 1}) &= \norm{\hat \theta - \bar{\theta}_{0}}_{P_{\theta}}^2 + \norm{\hat x_0 - \bar x_0}_{P_\text{x}}^2 + \nonumber \\
		& \quad +\sum_{j=0}^{L - 1} \norm{\hat{w}_j}_Q^2, \label{eq:FIE_objective}
\end{align}
where the weighting matrices $P_{\mathrm{x}}, P_{\theta}, Q \succ 0$ are tuning matrices.
A natural choice is the inverse of the covariance matrices of the initial state, parameter, and process/measurement noise, respectively.
An estimate of the system parameters is then obtained by solving the following nonlinear program offline
\begin{subequations}\label{eq:FIE}
\begin{alignat}{2}
\label{eq:FIE_cost}
\min_{\hat{\theta}, \{ \hat{x}_j \}, \{ \hat{w}_j \} } & V_{\mathrm{SysID}}(\hat{\theta}, \hat{x}_0, \{ \hat{w}_j \} )\\
\label{eq:FIE_1}
\text{s.t. } \hat{x}_{j+1} &= f(\hat{x}_{j},u_{j},\hat{w}_{j},\hat{\theta}),&&\ j\in\mathbb{I}_{[0, L - 1]},\\
\label{eq:FIE_2}
y_{j} &= h(\hat{x}_{j},u_{j},\hat{w}_{j},\hat{\theta}),&&\ j\in\mathbb{I}_{[0, L - 1]},\\
\label{eq:FIE_3}
\hat w_j & \in\mathbb{W}, \hat x_j \in \mathbb X_{\mathrm{ID}}, &&\ j\in\mathbb{I}_{[0, L - 1]}, \\
\label{eq:FIE_4}
\hat{\theta} &\in \Theta.
\end{alignat}
\end{subequations}
A (non-unique) minimizer of~\eqref{eq:FIE} is denoted as $\hat{\theta}^*$, $\{ \hat{x}_j \}^*$, $ \{ \hat w_j \}^*$ and the resulting parameter estimate as $\hat{\theta}=\hat{\theta}^*$.
Note that $\theta_{\mathrm{lh}}$ obtained above can be enforced within~\eqref{eq:FIE_4}.
The optimization problem~\eqref{eq:FIE} is non-convex, even for linear system dynamics.
Therefore, the solution is highly sensitive to the initialization of the solver.
To overcome this, it is essential to either run the optimization problem from multiple different initialization points or to warm-start the problem with a good initial guess, e.g., by first running a state estimator based on the prior parameter value $\bar{\theta}_0$ to initialize the sequence of states $\{ \hat x_j \}$.

%%%%%%%%%%%%%%%%%%%%%%%%%%%%%%%%%%%%%%%%%%%%%%%%%%%%%%%%%%%%%%%%%%%%%%%%%%%%%%%%
\section{STATE ESTIMATION}
\label{sec:estimation}

Given the discrete system dynamics \eqref{eq:dynamics_dt}, as well as the parameter estimate $\hat{\theta}$ obtained through system identification in Section~\ref{sec:sys_ID}, we introduce an online approach to obtain an estimate $\hat{x}_t$ of the system state at each time step $t$ given past input and output data $\{u,y\}$.
In particular, we rely on a moving horizon estimation (MHE) approach, which is an optimization-based estimation scheme that considers the past state estimate $\hat{x}_{t-M_t}$, obtained at time step $t-M_t$, as well as input and output data $\{u_j,y_j\}_{j=t-M_t}^{t-1}$ within a window $M_t$. To account for the initialization phase with limited data available, we define $M_t = \min\{t,M\}$, where $M\in\mathbb{I}_{\geq 0}$ is a fixed (bounded) horizon length.
The MHE approach optimizes over the initial state estimate $\hat{x}_{t-M_t|t}$ 
and a sequence of $M_t$ noise estimates $\hat{w}_{\cdot|t}=\left\{\hat{w}_{j|t}\right\}_{j=t-M_t}^{t-1}$ and output estimates $\hat{y}_{\cdot|t}=\left\{\hat{y}_{j|t}\right\}_{j=t-M_t}^{t-1}$.
Given  $\hat{\theta}$, $\hat{x}_{t-M_t|t}$ as well as $\hat{w}_{\cdot|t}$, the sequence of state estimates along the horizon can be computed from the system dynamics~\eqref{eq:dynamics_dt_1}.
The objective of the MHE problem is chosen as
\begin{align}
		&V_{\mathrm{MHE}}(\hat{x}_{t-M_t|t},\hat{w}_{\cdot|t},\hat{y}_{\cdot|t},t) = \eta^{M_t}\|\hat{x}_{t-M_t|t}-\hat{x}_{t-M_t}\|_{P}^2 \nonumber \\
		& \quad +\sum_{j=1}^{M_t}\eta^{j-1}\left(\|\hat{w}_{t-j|t}\|_Q^2+\|\hat{y}_{t-j|t}-y_{t-j}\|_R^2 \right), \label{eq:MHE_objective}
\end{align}
where $P$, $Q$ and $R$ are appropriate covariance matrices. 
The discount factor $\eta \in (0,1)$ ensures that more recent measurements have a greater impact on the optimization problem. 
The state estimate at time step $t$ is then obtained by solving the following nonlinear program (NLP)
\begin{subequations}\label{eq:MHE_IOSS}
\begin{alignat}{2}\label{eq:MHE_IOSS_cost}
\min_{\hat{x}_{t-M_t|t},\hat{w}_{\cdot|t}}&V_{\mathrm{MHE}}(\hat{x}_{t-M_t|t},\hat{w}_{\cdot|t},\hat{y}_{\cdot|t},t) \\ \label{eq:MHE_IOSS_1}
\text{s.t. }
\hat{x}_{j+1|t} &=f(\hat{x}_{j|t},u_{j},\hat{w}_{j|t},\hat{\theta}), &&\hspace{-3mm} j\in\mathbb{I}_{[t-M_t,t-1]},\\
\label{eq:MHE_IOSS_2}
\hspace{1.05cm} \hat{y}_{j|t}&=h(\hat{x}_{j|t},u_{j},\hat{w}_{j|t},\hat{\theta}), &&\hspace{-3mm}j\in\mathbb{I}_{[t-M_t,t-1]},\\
\label{eq:MHE_IOSS_3}
\hspace{1.05cm}\hat{w}_{j|t}&\in\mathbb{W},\ \hat{y}_{j|t}\in\mathbb{Y}, &&\hspace{-3mm}j\in\mathbb{I}_{[t-M_t,t-1]}, \\
\label{eq:MHE_IOSS_4}
\hspace{1.05cm}\hat{x}_{j|t}&\in\mathbb{X}, &&\hspace{-3mm}j\in\mathbb{I}_{[t-M_t,t]},
\end{alignat}
\end{subequations}
where \eqref{eq:MHE_IOSS_1}, \eqref{eq:MHE_IOSS_2} are the measurement and dynamic model constraints, \eqref{eq:MHE_IOSS_3} refers to noise and output bounds and \eqref{eq:MHE_IOSS_4} are state constraints.
A (non-unique) minimizer of~\eqref{eq:MHE_IOSS} is denoted as $\hat{x}_{t-M_t|t}^*$, $\hat{w}_{\cdot|t}^*$, and the resulting state estimate as
\begin{align}\label{eq:state_estimate}
    \hat{x}_t = \hat{x}_{t|t}^*.
\end{align}
Note that a particular benefit of MHE compared to other nonlinear estimation approaches, e.g., the classical EKF~\cite{McGee1985}, in the context of safety-critical applications is the ability to establish theoretical properties for the resulting state estimate~\eqref{eq:state_estimate}. 
In particular, if the true parameters are obtained during system identification, the system~\eqref{eq:dynamics_dt} is detectable~\cite[Ass.~1]{Schiller2023}, and $\eta < 1$ and the horizon length $M$ are chosen sufficiently large, then the estimation error, i.e., $x_t - \hat{x}_t$, resulting from the MHE approach~\eqref{eq:MHE_IOSS} is robustly stable~\cite[Cor.~1]{Schiller2023}, which implies that it is upper bounded at all time steps by decaying terms involving the initial state estimation error and the noise acting on the system.
For bounded uncertainties, this allows to robustify a control algorithm to state estimation errors, compare~\cite{koehler2021Output}.
If the true system parameters are not recovered during system identification, MHE allows to obtain stable state estimates even if the available online data is not sufficiently informative~\cite{Muntwiler2023}. Additionally, states and parameters can be estimated jointly when the data is informative~\cite{schiller2023moving}.

%%%%%%%%%%%%%%%%%%%%%%%%%%%%%%%%%%%%%%%%%%%%%%%%%%%%%%%%%%%%%%%%%%%%%%%%%%%%%%%%
\section{CONTROL}\label{sec:control}

In the following, we first provide an overview of a high-level control scheme to compute a control input based on the current state estimate $\hat{x}_t$, and then introduce the low-level controllers onboard the car.

\subsection{Model Predictive Contouring Control}
\label{sec:mpcc}

Model predictive contouring control (MPCC) has been used extensively in planning and control for autonomous racing \cite{Carron2023, Liniger2015RCCars, Froehlich2022}. It takes into account the track boundaries as constraints and optimizes the progress along a reference path while trading off path following and performance. It can be formulated as the NLP
\begin{subequations}\label{eq:mpcc}
    \begin{alignat}{3}
        \label{eq:mpcc-cost}
        \min_{\{u_i\}_{i=0}^{i=N}} &\sum_{i=0}^{N}\   \| \varepsilon(x_i) \|_Q^2 - Q_{\mathrm{adv}} \gamma(x_i) + \| u_i \|_R^2 \\
        \label{eq:mpcc-dynamics}
        \text{s.t. } & x_{i+1} = f(x_i, u_i, 0, \hat{\theta}), \ x_0 = \hat{x}_t, && \hspace{-5mm}i\in\mathbb{I}_{[0,N]},\\
        \label{eq:mpcc-constraints}
        & x_i \in \mathbb{X}, \; u_i \in \mathbb{U}, \gamma(x_i) > 0,  && \hspace{-5mm}i\in\mathbb{I}_{[0,N]},
    \end{alignat}
\end{subequations}
where $\varepsilon(x) = \begin{bmatrix} \varepsilon_l(x), \varepsilon_c(x) \end{bmatrix}^\top$ denotes the lag and contour error, defined as longitudinal and lateral error from the reference trajectory \cite{Carron2023}. The matrices $Q \in \mathbb R^2$, $Q_\text{adv} \in \mathbb R$, and $R \in \mathbb R^2$ are tuning parameters. The function $\gamma(x)$ describes the progress along the reference path. State and input constraints are specified by the sets $\mathbb X$ and $\mathbb U$. The MPCC problem~\eqref{eq:mpcc} is initialized with the state estimate~$\hat{x}_t$ obtained from the MHE~\eqref{eq:MHE_IOSS}, while the model parameters~$\hat \theta$ are found through the methods from~\cref{sec:sys_ID}. Note that~\eqref{eq:mpcc} constitutes a nominal MPCC for system~\eqref{eq:dynamics_dt}, which has proven to be inherently robust to small model perturbations~\cite{numerow2024inherently}.

\subsection{Low-level Control}

Onboard Chronos, low-level controllers track the reference given by the solution of~\eqref{eq:mpcc}. The input to the low-level controller is either the pair $\{ \delta, T \}$ or $\{ \delta, v_{\mathrm x} \}$, where $\delta$ is the steering angle and $T$ the input torque; as an alternative, the reference longitudinal velocity $v_{\mathrm x}$ can be specified. All low-level controllers run at \SI{250}{\hertz}.

\subsubsection{Steering Angle Control} Given a map from the onboard steering potentiometer's voltage $V_\delta$ to the resulting steering angle $\delta(V_\delta)$, we estimate the current steering angle and track it using a PID controller that outputs a PWM signal to the motor chip controller of the steer DC motor.

\subsubsection{Longitudinal Velocity Control}

From the wheel encoder sensor model~\eqref{eq:wheel-encoder-model}, we can obtain an onboard estimate of the longitudinal velocity from the rear wheel velocities as $v_x = (\Omega_\mathrm{rl} + \Omega_\mathrm{rr}) / (2 \cdot r)$. A PID controller closes the loop to track a given reference velocity.

%%%%%%%%%%%%%%%%%%%%%%%%%%%%%%%%%%%%%%%%%%%%%%%%%%%%%%%%%%%%%%%%%%%%%%%%%%%%%%%%

\input{sections/hardware-experiments}

%%%%%%%%%%%%%%%%%%%%%%%%%%%%%%%%%%%%%%%%%%%%%%%%%%%%%%%%%%%%%%%%%%%%%%%%%%%%%%%%
\section{CONCLUSIONS}

In this work, we present a low-cost and readily available sensor setup and optimization pipeline for miniature car-like robots where hardware, firmware, and software have been open-sourced.
Our method allows for optimization-based system identification, state estimation (MHE), and controls (MPCC).
To this end, we introduce analytical sensor models for the Lighthouse positioning system, inertial measurement unit, and wheel encoders, as well as a modified bicycle dynamic model. 
Finally, we validate and demonstrate that our setup allows for real-time control and estimation by deploying it on miniature race cars. 
%%%%%%%%%%%%%%%%%%%%%%%%%%%%%%%%%%%%%%%%%%%%%%%%%%%%%%%%%%%%%%%%%%%%%%%%%%%%%%%%
\section*{ACKNOWLEDGMENTS}
\noindent We would like to thank Karl Marcus Aaltonen, Jerome Sieber, Shengjie Hu, Joshua Naef, Maria Krinner, Griffin Norris, Moritz H\"usser, Marvin Harms, and Andrea Zanelli for the support with the development of Chronos and~CRS.

%%%%%%%%%%%%%%%%%%%%%%%%%%%%%%%%%%%%%%%%%%%%%%%%%%%%%%%%%%%%%%%%%%%%%%%%%%%%%%%%

\bibliographystyle{ieeetr}
\bibliography{bibliography}

\addtolength{\textheight}{-3cm}   % This command serves to balance the column lengths
\ifarXiv
\input{sections/appendix}
\fi
\end{document}

%% file: sections/introduction.tex
\vspace{0.1cm}
\noindent {\small {\bf Code}: \url{https://gitlab.ethz.ch/ics/crs}}

\noindent {\small {\bf Dataset}: \url{https://gitlab.ethz.ch/ics/crs/-/tree/main/datasets}}

\noindent {\small {\bf Video}: \url{https://youtu.be/vKF30Sol8Qk}}

\section{INTRODUCTION}

Experimental work plays a crucial role in assessing the effectiveness and limitations of control and estimation methods. However, the execution of hardware experiments faces various challenges, even for relatively simple setups. These challenges include the high costs associated with procuring and setting up hardware platforms, the difficulty in identifying models and their parameters needed for model-based state estimation and control, the complexity involved in developing the control software architecture, and the scarcity of (advanced) open-source control and estimation algorithms~\cite{zhou2023impact, Pineau2021Improving, How2018Control} that could be reused to speed up the deployment. 
To mitigate these limitations, we present an indoor low-cost localization\footnote{By localization, we refer to estimating the configuration state (position and orientation) of a robotic system in three-dimensional space, while state estimation refers to estimating the full system state, e.g., including velocities and yaw rates.} system, an optimization-based approach for system identification, and a moving horizon estimation (MHE) approach for an updated version of the miniature, low-cost car-like robot called Chronos~\cite{Carron2023}. 

\subsubsection*{Contribution}
The contributions of this paper include:
\begin{enumerate}[wide, labelwidth=!, labelindent=0pt]
    \item The introduction of an open-source and low-cost hardware platform (Section~\ref{sec:hw}), enhancing the Chronos car introduced in~\cite{Carron2023} with custom-built wheel encoders and the off-the-shelf Lighthouse positioning deck~\cite{Taffanel2021}, see Fig.~\ref{fig:overview}. The presented robotics platform costs less than \$\,700.
    Fusing the information of the different sensors allows us to obtain accurate estimates of the state of Chronos.
    The presented hardware setup significantly lowers the bar for hardware experiments, which is paramount to testing algorithms in realistic settings in the control, robotics, and machine learning research and education communities.
    The presented hardware (electronic schematics and printed circuit board (PCB) designs), software, and firmware are open-source under a BSD 2-clause license.
    The platform is designed modularly, allowing for different system models, e.g., model rockets~\cite{Spannagl2021}, and multi-agent applications, e.g., coverage control~\cite{Rickenbach2024}.
    \item Improvements of the car modelling (Section~\ref{sec:modelling}). In particular, we present an extension of the standard bicycle model by splitting the motor force to the front and rear axle such that one formulation can be used to model front-, rear-, and all-wheel drive cars.
    In addition, an approximation for the Pacejka tire forces is introduced to overcome singularities at zero longitudinal velocities.
    \item A complete pipeline for advanced optimization-based system identification (Section~\ref{sec:sys_ID}), state estimation (Section~\ref{sec:estimation}), and control (Section~\ref{sec:control}).
    In many practical applications, system identification and state estimation are limiting factors for advanced control design, e.g., due to the nonlinear nature of the considered systems and because the system state can not be measured directly.
    The presented pipeline of optimization-based algorithms aims to support the design and application of model learning and control with guarantees, e.g., stability and constraint satisfaction. An overview of the provided software frame work is given in \ifarXiv\cref{sec:appendix-crs-overview}\else\cite[Appendix~A]{Bodmer2024} \fi.
    \item Extensive hardware experiments (Section~\ref{sec:experiments})  show that the available low-cost sensors allow for accurate state estimation and for controlling the system in closed-loop, even in the presence of sensor failure for limited time intervals. Furthermore, we make the dataset used for system identification and the open-loop experiments available online. We believe that these datasets can become a useful benchmark for nonlinear system identification and estimation. 
\end{enumerate}

\subsubsection*{Related Work}
Localization systems are crucial in enabling autonomous navigation for various robotic platforms. Possible approaches include motion capture, overhead camera systems, and onboard sensors. Motion capture systems, used in platforms like~\cite{Pickem2017Robotarium, Chalaki2022IDS3C, Buckman2022MiniCity, Hyldmar2019Cambridge} offer unparalleled accuracy and precision. However, their widespread adoption is hindered by their high cost, making them impractical for low-cost applications.
On the other hand, overhead camera systems, such as those employed in~\cite{Kloock2021CPM, Wilson2016Pheeno, Dong2023MCCT, Liniger2015RCCars, Carron2013Lego} offer a more cost-effective solution. However, their accuracy is typically lower and the software is not open-source, further limiting their accessibility and customization for specific applications.
Alternatively, onboard sensor-based localization systems, like those utilized in~\cite{Paull2017Duckietown, okelly2019f110}, offer a more versatile solution that does not require external infrastructure. However, they may not always be suitable for applications with stringent size limitations. 

Car-like robots are often modeled using a bicycle model, with tire forces modeled with the Pacejka model~\cite{Carron2023, rajamani2011vehicle}.
A modified model to account for all-wheel drive (AWD) configurations was introduced in~\cite{Raji2022}.
Models obtained from first principles usually contain parameters that cannot be measured directly and, therefore, require parametric system identification methods relying on input/output data only.  In general, these methods are scarce for nonlinear systems.
Promising methods make use of (non-convex) optimization approaches to find sequences of system states and parameters which maximize the likelihood of available input/output data, compare, e.g.,~\cite{bock1983recent,valluru2017development,simpson2023efficient}.
Similarly, sensor calibration can be performed in an optimization-based manner, as, e.g., done in~\cite{Taffanel2021} for the Lighthouse positioning system.

In practical control applications, state estimation plays a crucial role. 
Compared to the widely used extended Kalman filter (EKF)~\cite{McGee1985}, an MHE approach~\cite[Chap.~4]{rawlings2020model} is very promising for safety critical systems, as it does not rely on linearization of the system model and can provide robust stability of the resulting state estimate~\cite{Schiller2023}.
Outliers can be rejected within an MHE by neglecting the corresponding measurements~\cite{alessandri2016moving} or using an appropriate objective function~\cite{Gharbi2021}.
System parameters that cannot be identified perfectly offline or are slowly time-varying (e.g., due to tire wear), can be jointly estimated with system states in an MHE if the online measurements are informative enough~\cite{schiller2023moving}, while robustly stable state estimates can be obtained even if the data is not informative~\cite{Muntwiler2023}.
While MHE has been applied for (offline) estimation of car positions based on real-world vehicle test data~\cite{Brembeck2019} and joint state and friction estimation in simulation~\cite{zanon2013nonlinear}, applications of MHE to real-world dynamical systems are rare.

\subsubsection*{Notation}

We use $\mathbb I_{[a, b]}$ to denote the set of integers $i \in \mathbb N$ with $a \leq i \leq b$ and $\mathbb I_{\ge a}$ to denote the set of integers greater or equal to $a$. The sequence $(x_0, \ldots, x_N)$ is denoted as $\{ x_j \}_{j = 0}^N$, or abbreviated as $\{ x_j \}$. The weighted square norm of $x$ is $\norm{x}_Q^2 = x^\top Q x$. $R_{\chi}(\varphi) \in \operatorname{SO}(3)$ are elementary rotation matrices of angle $\varphi$ around the coordinate axis $\chi$.

%% file: sections/hardware.tex
\section{CHRONOS ONBOARD AND EXTERNAL SENSORS}\label{sec:hw}

In the following, we provide a brief hardware overview of the onboard sensors mounted on the Chronos car~\cite{Carron2023} and the external Lighthouse positioning system. For an in-depth overview of the CRS software framework, see 
\ifarXiv\cref{sec:appendix-crs-overview}\else\cite[Appendix~A]{Bodmer2024} \fi
\,and \cite{Carron2023}.

\subsection{Inertial Measurement Unit}

The car is equipped with an inertial measurement unit (IMU) aligned with the body axes. It provides linear acceleration and angular velocity measurements at a rate of~\SI{250}{\hertz}.

\subsection{Wheel Encoders}

Using rotary encoders as a source of odometry is an established practice in robotics \cite{yurtsever2020survey}, and the twist estimates from an encoder system can be used in dead reckoning scenarios.
To obtain an estimate of the angular velocities of the wheels $\Omega_i$, we mount eight small magnets in the wheel rims and a small PCB with a hall effect sensor close to each wheel axle. We sample the wheel speeds at \SI{250}{\hertz}. 

\setcounter{figure}{1}
\begin{figure}[t]
    \centering
    \includegraphics[width=1\columnwidth]{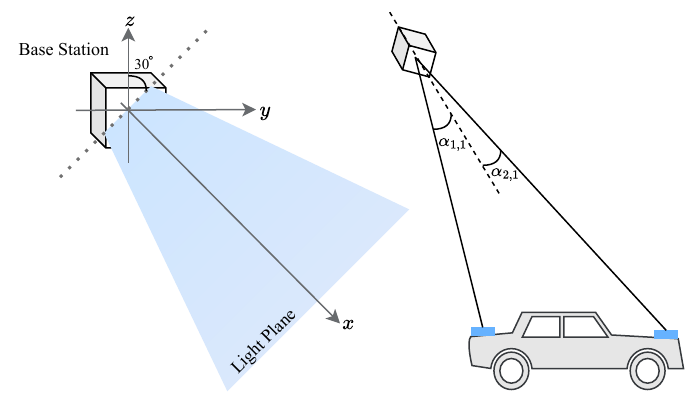}
    \caption{The left figure demonstrates a single sweep of the light plane by the base station. Both light planes rotate around the $z$-axis. The second light plane is rotated by \SI{60}{\degree} compared to the first plane (depicted here in blue). The right image demonstrates the angles measured by the onboard sensors. The angles $\alpha_{1,1}$, $\alpha_{2,1}$ refer to the angle measurement of the first and second sensor for the first light plane sweep.}
    \label{fig:lighthouse-working-principle}
    \vspace{-15px}
\end{figure}

\subsection{Lighthouse Positioning System}

The Lighthouse positioning system consists of (potentially) multiple base stations, each emitting two rotating light planes, and photodiode sensors mounted on an object to be tracked. The base stations modulate a data stream onto the emitted light planes, synchronized with their rotation. Using a third-party lightweight sensor board called Lighthouse deck, distributed by Bitcraze~\cite{Taffanel2021}, the demodulated data stream can be used to determine the angle~$\alpha$ of the light plane relative to the base station when the four sensors on the deck are hit. Through triangulation, the position of the sensors relative to the base station can be determined. Fig.~\ref{fig:lighthouse-working-principle} depicts the system's working principle.

We use version 2.0 of the base stations. In the planar 2D tracking case, a single base station above the racetrack is sufficient to cover an area of around \SI{6}{\meter} by \SI{6}{\meter} and provides updates at around \SI{50}{\hertz}. Additional base stations can be used to cover larger areas and increase the update rate.

%% file: figures/tikz/dynamic.tex
\usetikzlibrary{calc}
\begin{tikzpicture}
	\tikzset {
		wheel/.style n args={3}{
			rectangle,
			fill,
			black,
			minimum width={#1},
			minimum height={#2},
			rotate={#3},
			rounded corners,
	}};
    \tikzset {
		graywheel/.style n args={3}{
			rectangle,
			fill,
		      gray,
			minimum width={#1},
			minimum height={#2},
			rotate={#3},
			rounded corners,
	}};
	\def\height{4cm};
	\def\radius{\height/5};
	\def\width{\height/15};
    \def\carwidth{2cm}
	\def\cor{\height/1.2};
	\def\steer{30};
	\def\bet{atan(\height/2/(\cor))};
	\def\arcradius{0.7};
	\def\yaw{30};	
	\def\lengthsep{1.5};
	\def\dynang{(\bet)};	
	
	\begin{scope}[rotate=\yaw]

    \node[graywheel={\radius}{\width}{\yaw}] (wrl) at (-\height/2,\carwidth/2){};
    \node[graywheel={\radius}{\width}{\yaw}] (wrr) at (-\height/2,-\carwidth/2){};
    \node[graywheel={\radius}{\width}{\steer+\yaw}] (wfl) at (\height/2,\carwidth/2){};
    \node[graywheel={\radius}{\width}{\steer+\yaw}] (wfr) at (\height/2,-\carwidth/2){};

    \draw[gray] (wrr) -- (wrl);
    \draw[gray] (wfr) -- (wfl);
 
	\node[wheel={\radius}{\width}{\yaw}] (wr) at (-\height/2,0){};
	\node[wheel={\radius}{\width}{\steer+\yaw}] (wf) at (\height/2,0){};
	\node[draw,inner sep=\height/50, circle,radius = \height/50,fill] (cog) at (0,0){};
	
	\draw[dashed] (wf) -- ++({\steer}:1);
	\draw[dashed] (wf) -- ++(1,0);
	\draw[->] (wf.center) ++(0:\arcradius) arc (0:\steer:\arcradius) node[midway,shift={(0.15,0.15)}]{$\delta$};
	
	\def\vell{1.5};	
	\draw (wr) -- (wf);
	\draw[->,blue,thick] (cog) -- ++({\bet}:\vell) node()[shift={(-0.2,0)}]{$v$};
	\draw[->] (cog) ++(0:\arcradius) arc (0:{\bet}:\arcradius) node[anchor = west,shift={(0.1,0)}]{$\beta$};
	
	\draw[->,blue,thick] (cog) -- ++(0,{sin(31)*\vell} ) node()[shift={(-0.1,-0.4)}]{$v_{\mathrm{y}}$};
	\draw[->,blue,thick] (cog) -- ++({cos(31)*\vell},0 ) node()[shift={(0,0.2)}]{$v_{\mathrm{x}}$};
    \draw[->,red,thick] (cog) -- ++(-{cos(31)*\vell*0.5},0 ) node()[shift={(0.3,-0.1)}]{$F_{\mathrm{fr}}$};
	
	\draw[->,blue,thick] (wr.center) -- ++({\dynang}:1);	
	\draw[->] (wr.center) ++(0:\arcradius) arc (0:\dynang:\arcradius) node()[shift={(0.4,0.1)}]{\footnotesize $\alpha_{\mathrm{r}}$};
	
	\draw[->,blue,thick] (wf.center) -- ++({\dynang+\steer}:1);
	\draw[->] (wf.center) ++(\steer:\arcradius) arc (\steer:{\dynang+\steer}:\arcradius) node()[shift={(0.25,0.1)}]{\footnotesize $\alpha_{\mathrm{f}}$};

	\draw[->,red,thick] (wr) -- node()[anchor = east]{$F_{\mathrm{y,r}}$} ++(0,\arcradius);
	\draw[<-,red,thick] (wr) -- node()[anchor = north]{$F_{\mathrm{x,r}}$} ++(-1.5*\arcradius,0);
	\draw[->,red,thick] (wf) -- node()[shift={(0.07,0.25)}]{$F_{\mathrm{y,f}}$} ++({\steer+90}:\arcradius);
	\draw[<-,red,thick] (wf) -- node()[anchor = west, shift={(-0.1,-0.1)}]{$F_{\mathrm{x,f}}$} ++({\steer}:{-1.5*\arcradius});

    \draw[->,blue,thick] (wrr) -- node()[anchor = north, shift={(0.2,0)}]{$v_{\mathrm{w,rr}}$} ++(1.5*\arcradius,0);
    \draw[->,blue,thick] (wrl) -- node()[anchor = north, shift={(0.4,0.15)}]{$v_{\mathrm{w,rl}}$} ++(1.5*\arcradius,0);

    \draw[->,blue,thick] (wfl) -- node()[anchor = west, shift={(-0.1,-0.1)}]{$v_{\mathrm{w,fl}}$} ++({\steer}:{1.5*\arcradius});
    \draw[->,blue,thick] (wfr) -- node()[anchor = west, shift={(-0.1,-0.1)}]{$v_{\mathrm{w,fr}}$} ++({\steer}:{1.5*\arcradius});

	\draw[|-|] ($(wr)+(0,\lengthsep)$) -- node()[anchor = north]{$l_{\mathrm{r}}$} ($(cog)+(0,\lengthsep)$);
	\draw[|-|] ($(cog)+(0,\lengthsep)$) -- node()[anchor = north]{$l_{\mathrm{f}}$} ($(wf)+(0,\lengthsep)$);

    \draw[|-|] ($(wfr)+(\lengthsep,0)$) -- node()[anchor = south west]{$b_{\mathrm{car}}$} ($(wfl)+(\lengthsep,0)$);

    \draw[|-|] ($(wr)+(-\radius/2,-0.9*\lengthsep)$) -- node()[anchor = north, shift={(0.4,0.1)}]{$2\cdot r_{\mathrm{w}}$} ($(wr)+(\radius/2,-0.9*\lengthsep)$);
	
	\def\psiangle{60};
	\draw[->] (cog.center) ++(\psiangle:1) arc (\psiangle:180-\psiangle:1) node[midway,shift={(0,0.3)}]{$\omega$};
	\end{scope}
	
	\draw[dashed] (cog.center) -- ($(cog.center)+(1,0)$);
	\draw[->] (cog) ++(0:\arcradius) arc (0:\yaw:\arcradius) node[shift={(0.3,-0.1)}]{$\psi$};
	
	\def\loff{3.5cm};
	\def\doff{2.7cm};
	\draw[->] (-\loff,-\doff) -- (\loff,-\doff) node()[anchor = south]{$x$};
	\draw[->] (-\loff,-\doff) -- (-\loff,\doff) node()[anchor = east]{$y$};
	\draw[densely dotted] (cog) -- (0,-\doff - 0.2cm) node[anchor = north]{$x_{\mathrm{p}}$};
	\draw[densely dotted] (cog) -- (-\loff - 0.2cm,0) node[anchor = east]{$y_{\mathrm{p}}$};

\end{tikzpicture}

%% file: sections/hardware-experiments.tex
\section{HARDWARE EXPERIMENTS}\label{sec:experiments}

This section presents the results of real-world experiments performed on the hardware described in \cref{sec:hw}. 
The Lighthouse base station was mounted approximately over the center of the track on a tripod, with the base station pointing straight down to cover the largest area of the track. Where applicable, we use Qualisys, a high-quality motion capture system, as a ground-truth reference.
Note that all relevant parameters can be found in the code available online.

\subsection{Lighthouse Calibration}

For calibration, we are interested in two key figures: what is the expected accuracy of the Lighthouse positioning system in static scenarios and how many point-angle correspondences are needed to achieve a sufficient estimate of~$\theta_{\mathrm{lh}}$. We use the calibration procedure detailed in Section \ref{sec:lh_calibration} and collect angle measurements in 24 static locations. We then solve~\eqref{eq:lighthouse-calibration-cost} for increasing point-angle correspondences and calculate the resulting residual positions.

\begin{figure}
    \centering
    \includegraphics[width=.8\linewidth]{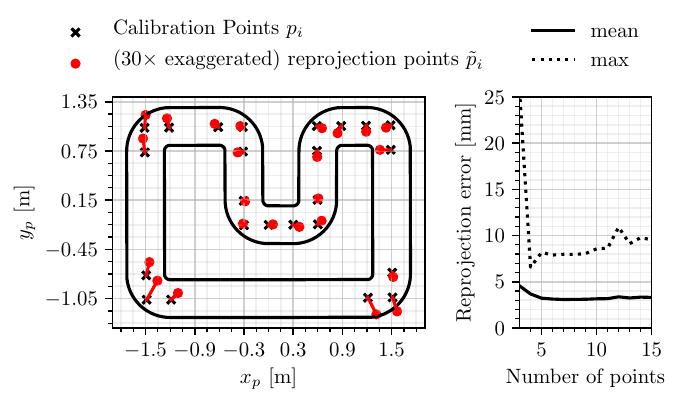}
    \caption{Left: Reprojection error (scaled $30\times$) of calibration points. Right: Mean/maximum calibration residual by number of points.}
    \label{fig:calibration-results}
    \vspace{-15px}
\end{figure}

As a measure of calibration fidelity, the angles measured during calibration are reprojected onto the plane using the crossing beams method detailed in \cite{Taffanel2021}, resulting in residuals with respect to the ground truth calibration points~$p_i$. \cref{fig:calibration-results} shows the reprojection errors displayed on the calibration points, where the error has been scaled 30\nobreakdash-fold. Also, mean and maximum residuals are evaluated for a varying number of calibration points~$n_\text{cal}$, arranged so that the density between points increases approximately equally when adding points. We find that by using only 5\nobreakdash--6 well-distributed points, a mean residual of \SI{3}{\milli\metre} can be achieved in static conditions. A small number of points may have residuals up to \SI{11}{\milli\metre}, in areas where the angle averaging described in \cref{sec:lh_calibration} yields a worse approximation, particularly in shallow angles. Note that the lower bound for achievable residuals here is given by the motion capture system's \SI{2}{\milli\metre}--\SI{3}{\milli\metre} residuals for the (ground-truth) calibration points.

\subsection{Optimization-based System Identification}

We implement the optimization problem~\eqref{eq:FIE} using CasADi~\cite{andersson2019casadi} and solve it using the interior point solver Ipopt~\cite{Waechter2005}. Using a model-free controller (PID) and an EKF, data is recorded as the car drives around the track. Based on the recorded control inputs and measurements we perform system identification. The prior parameter estimate $\bar \theta_0$ is chosen from estimates available for a related car model.

\begin{figure}
    \centering
    \includegraphics[width=.9\linewidth]{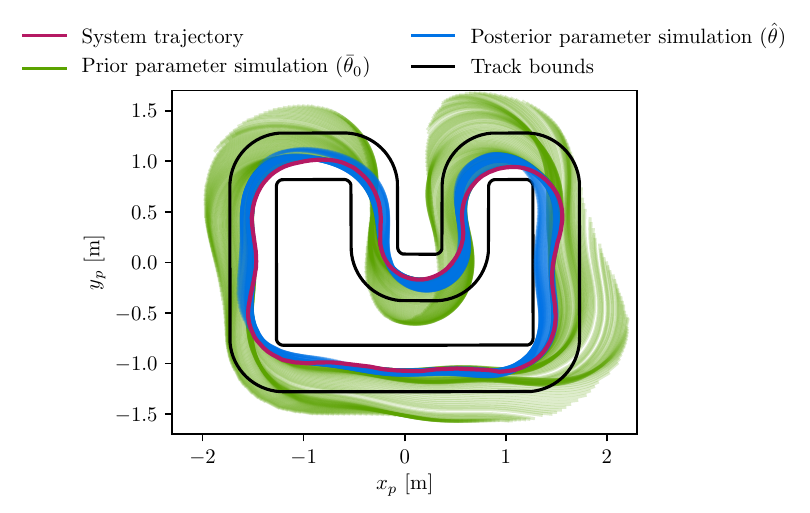}
    \captionof{figure}{Open-loop predicted trajectories from inputs $\{\tilde u_j \}_{j=t}^{t+M}$ based on the prior parameter estimate $\bar{\theta}_0$ and posterior identified system model $\hat \theta$.}
    \label{fig:sysid_openloop_comparison}
    \vspace{-13px}
\end{figure}

From the solution of~\eqref{eq:FIE}, $\hat \theta$ is obtained. We then predict open-loop trajectories $\tilde x(t; \theta)$, for both $\bar \theta_0$ and $\hat \theta$ on an unseen validation dataset $\{ \tilde u_j, \tilde y_j \}_{j = 0}^{L - 1}$ by simulating~\eqref{eq:dynamics_dt} over a fixed horizon $M$ for $j \in \mathbb I_{[0, L - M - 1]}$. In~\cref{fig:sysid_openloop_comparison}, the contrast between the open-loop predictions of the prior model, parametrized by $\bar \theta_0$, to the model using the identified parameters $\hat \theta$ is shown.  As an improvement metric, we compare to the measured system trajectory. We achieve a reduction in root-mean-squared error (RMSE) from \SI{0.24}{\meter} to \SI{0.09}{\meter} over an open loop prediction horizon of \SI{2}{\second}.

\subsection{Estimation Analysis}

To evaluate the effect of various sensors for the different estimation architectures, an EKF and MHE were implemented. For the MHE, we used Acados~\cite{verschueren2022acados} as a solver. 
The respective EKF and MHE RMSE for each state\footnote{Note that we do not provide the RMSE in yaw rates because we do not have accurate estimates of the ground truth for this state.} averaged over one run of twenty seconds are reported in Table \ref{tab:combined_table}.

\begin{table}[h]
    \caption{RMSE and standard deviation for different estimators}
    \centering
    \setlength{\tabcolsep}{2pt}% Shrink \tabcolsep by 30%
    \centering
    \begin{tabular}{l | c c c|c c c c c }
    \toprule
      \multirow{2}{*}{Type} & \multicolumn{3}{c|}{Sensors} & \multirow{2}{*}{$x$ \tiny[mm]} & \multirow{2}{*}{$y$ \tiny[mm]}& \multirow{2}{*}{$\phi$ \tiny[mrad]} & \multirow{2}{*}{$v_\mathrm{x}$ \tiny[mm/s]}& \multirow{2}{*}{$v_\mathrm{y}$ \tiny[mm/s]} \\ 
     &LH &IMU &WE &  & & & & \\ \midrule
     EKF & \cmark & \xmark & \xmark & 31\tiny$\pm 0.03$ & 21\tiny$\pm 0.02$ & \textbf{42}\tiny$\pm 0.02$ & 69\tiny$\pm 0.07$ & 67\tiny$\pm 0.05$ \\
     MHE & \cmark & \xmark & \xmark & 29\tiny$\pm 0.03$ & 18\tiny$\pm 0.02$ & 71\tiny$\pm 0.05$ & 99\tiny$\pm 0.10$ & 83\tiny$\pm 0.08$ \\ \midrule
     EKF & \cmark & \cmark & \xmark & 25\tiny$\pm 0.01$ & 16\tiny$\pm 0.01$ & 44\tiny$\pm 0.02$ & 375\tiny$\pm 0.38$ & 290\tiny$\pm 0.24$ \\ 
     MHE & \cmark & \cmark & \xmark & \textbf{23}\tiny$\pm 0.02$ & \textbf{15}\tiny$\pm 0.01$ & 62\tiny$\pm 0.06$ & 262\tiny$\pm 0.26$ & 241\tiny$\pm 0.24$ \\ \midrule
     EKF & \cmark & \xmark & \cmark & 30\tiny$\pm 0.03$ & 20\tiny$\pm 0.02$ & 49\tiny$\pm 0.03$ & 31\tiny$\pm 0.03$ & 42\tiny$\pm 0.04$  \\ 
     MHE & \cmark & \xmark & \cmark & 32\tiny$\pm 0.03$ & 21\tiny$\pm 0.02$ & 74\tiny$\pm 0.04$ & 34\tiny$\pm 0.03$ & 44\tiny$\pm 0.04$  \\ \midrule
     EKF & \cmark & \cmark & \cmark & 30\tiny$\pm 0.03$ & 20\tiny$\pm 0.02$ & 45\tiny$\pm 0.02$ & 26\tiny$\pm 0.03$ & 36\tiny$\pm 0.04$ \\ 
     MHE & \cmark & \cmark & \cmark & 32\tiny$\pm 0.03$ & 21\tiny$\pm 0.02$ & 45\tiny$\pm 0.03$ & \textbf{25}\tiny$\pm 0.02$  & \textbf{32}\tiny$\pm 0.03$  \\ 
     \bottomrule
    \end{tabular}
    \label{tab:combined_table}
    \vspace{-11px}
\end{table}

\looseness=-1
The estimators are evaluated on open-loop data collected using a joystick controller to evaluate the performance of the MHE without the effects of closed-loop feedback. 
We collect ground-truth data for all states of the car using Qualisys and record all sensor readings (Lighthouse (LH), Wheel Encoders (WE), IMU).
Note that running Qualisys and the Lighthouse positioning system at the same time can result in decreased performance since the infrared flashes of the Qualisys motion capture system can interfere with the demodulation of the Lighthouse frames -- both systems occupy a similar spectrum.
To limit this effect, the Qualisys system frequency is reduced to \SI{35}{\hertz}.
The estimators are then evaluated in an offline setting but process the data in real-time to ensure successful deployment to hardware.
The RMSE is calculated using nearest neighbor interpolation based on the measurement times from the Qualisys system.

A closer analysis of the results in Table~\ref{tab:combined_table} shows that the MHE and EKF performed similarly. 
Interestingly, the sole addition of the IMU sensor leads to improved position estimates at the expense of less accurate body velocities. 
Including the wheel encoders (which provide a measurement of the forward body velocity) mitigates these effects. 
While the exact cause is unknown, we attribute this effect to the high noise levels and potential bias on the IMU measurements, which can lead to large velocity errors if these quantities are not directly observed. 
Note that generally, adding sensors tends to improve the estimator performance.
While MHE allows us to establish theoretical guarantees, investigating settings where MHE results in an increased estimation performance compared to, e.g., an EKF remains an interesting question for future work.

\subsection{Closed-loop Control}

\begin{figure}
    \centering
    \includegraphics[width=.72\linewidth]{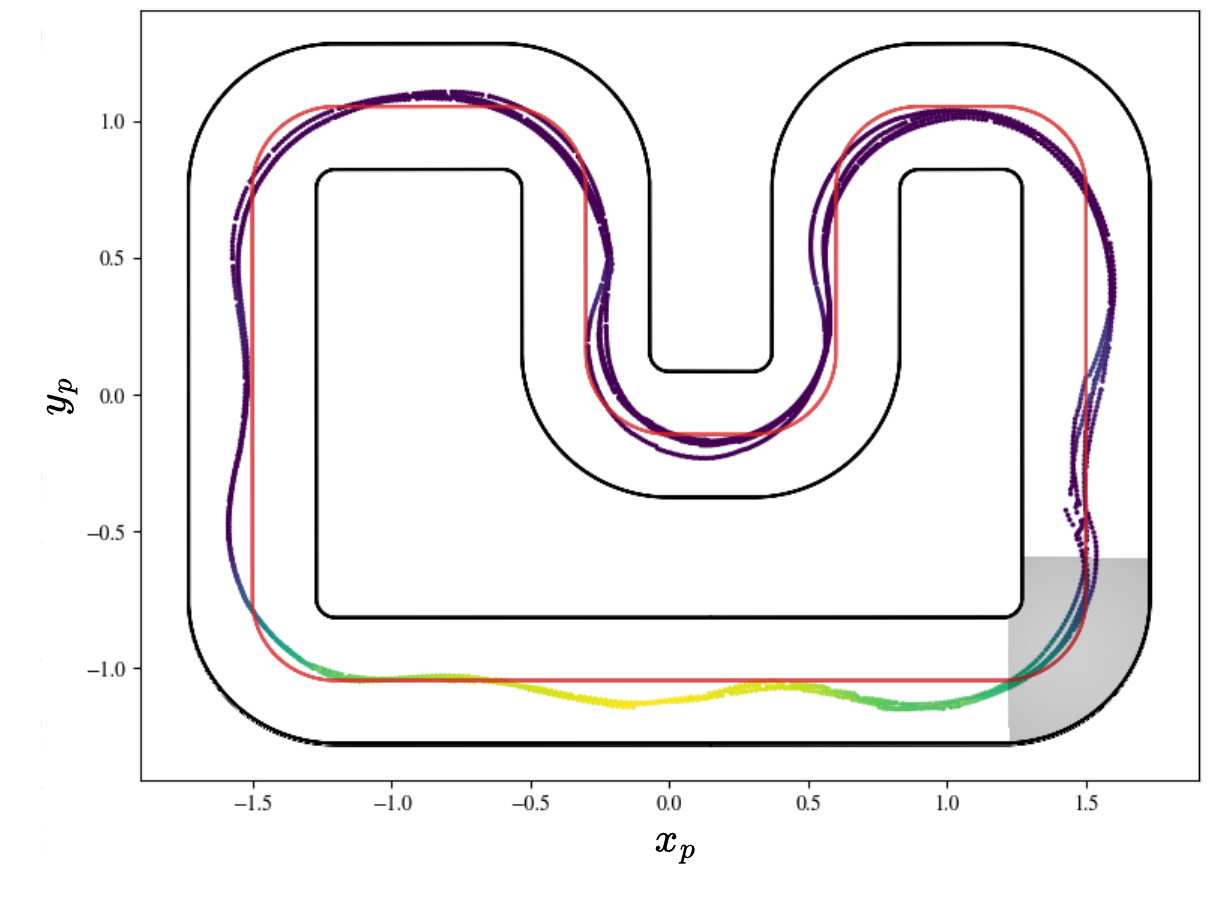}
    \caption{Closed-loop MPCC using the MHE with the Lighthouse positioning system, IMU, and wheel encoders. High velocities are shown in lighter colors and low velocities in darker colors, respectively. In the shaded gray area, no positioning measurements were available. The track centerline is shown in red. }
    \label{fig:closed_loop_mhe}
    \vspace{-18px}
\end{figure}

\looseness=-1
In this experiment, we achieve reliable small-scale autonomous racing with the proposed low-cost and readily available sensor setup.
The closed-loop experiments were performed using the MPCC and the MHE described in Section \ref{sec:control} and \ref{sec:estimation}, respectively. The MHE and MPCC both use a horizon length of $40$. The MHE is run at \SI{60}{\hertz}, while the MPCC is run at \SI{30}{\hertz}. \cref{fig:closed_loop_mhe} demonstrates the results of the closed-loop system using the Lighthouse positioning system, as well as the IMU and wheel encoders. Additionally, to verify the robustness of the closed-loop estimation and control architecture, we temporarily blinded the Lighthouse positioning system (\cref{fig:closed_loop_mhe} sections highlighted in grey) such that the MHE had to rely only on IMU and wheel encoder measurements. The Lighthouse positioning system measurements were missing for a maximum of \SI{1.127}{\sec} and \SI{0.753}{\sec} on average. At the MHE update rate of \SI{60}{\hertz}, positional information was missing for up to $68$ MHE calls. We note that the closed-loop system behavior was unaffected by these measurement dropouts and the car successfully completed the lap, showcasing the system can perform well even in
the presence of sensor failure for limited time intervals.

%% file: sections/appendix.tex
\clearpage
\appendix
\label{sec:appendix}

\subsection{Software, Hardware and Firmware Overview}
\label{sec:appendix-crs-overview}

In the following, we briefly summarize the main components of the CRS software, as well as the hardware and firmware of the Chronos car. For a more detailed introduction, we refer to \cite{Carron2023}. All of the software, hardware and firmware are released under a BSD~2-clause license.

\subsubsection{CRS framework}

\begin{figure*}[t]
    \centering
    \includegraphics[width=0.95\textwidth, page=1, trim={0 10cm 0 0}, clip]{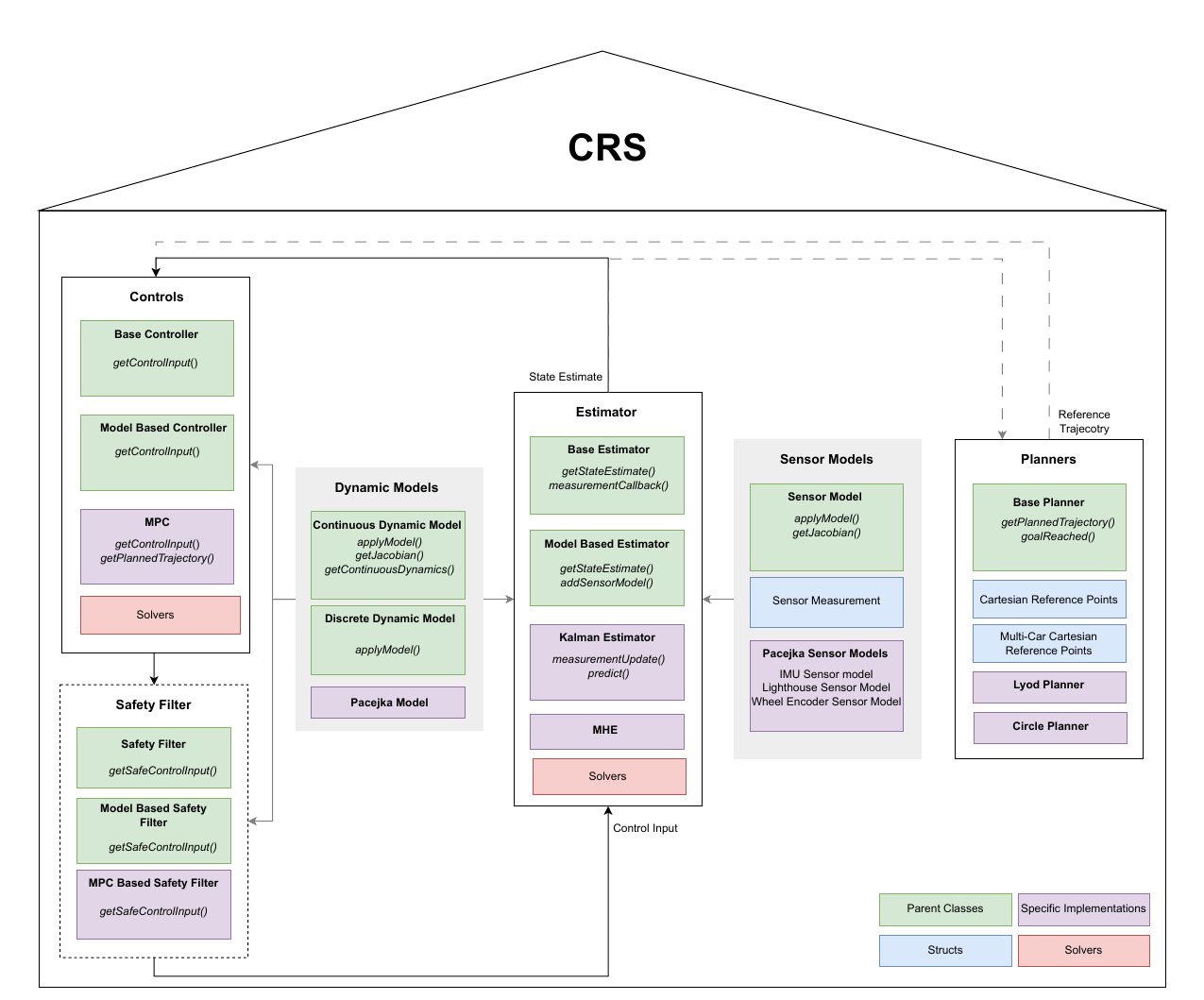}
    \caption{Overview of the CRS framework. The grey blocks (Dynamic Models and Sensor Models) are abstract interfaces of the model dynamics and measurement functions. The main control loop consists of the controller, estimator, and potentially a planner. The arrows show what information is shared among the main building blocks.}
    \label{fig:crs_overview}
    \vspace{-15px}
\end{figure*}

The CRS framework is based on ROS 1 and written primarily in C++, with some Python components. A split is emphasized between parts relying on ROS (\emph{ROS4CRS}) and those purely focused on control (\emph{CRS}), enabling modularity and encouraging reuse in other software. On a high level, \emph{CRS} offers a standard control pipeline, split into control, estimation, and navigation. \cref{fig:crs_overview} provides an overview of the main components implemented in the \emph{CRS} framework. Each of the main building blocks (controls, estimation, dynamic models, and sensor models) provides a base class (green blocks) from which specific implementations (purple blocks) can be derived. The arrows in \cref{fig:crs_overview} demonstrate the flow of information. Depending on the chosen implementation, the dashed lines visualize additional information.

\begin{figure*}[t]
    \centering
    \includegraphics[width=0.85\textwidth, page=2, trim={0 9cm 0 0}, clip]{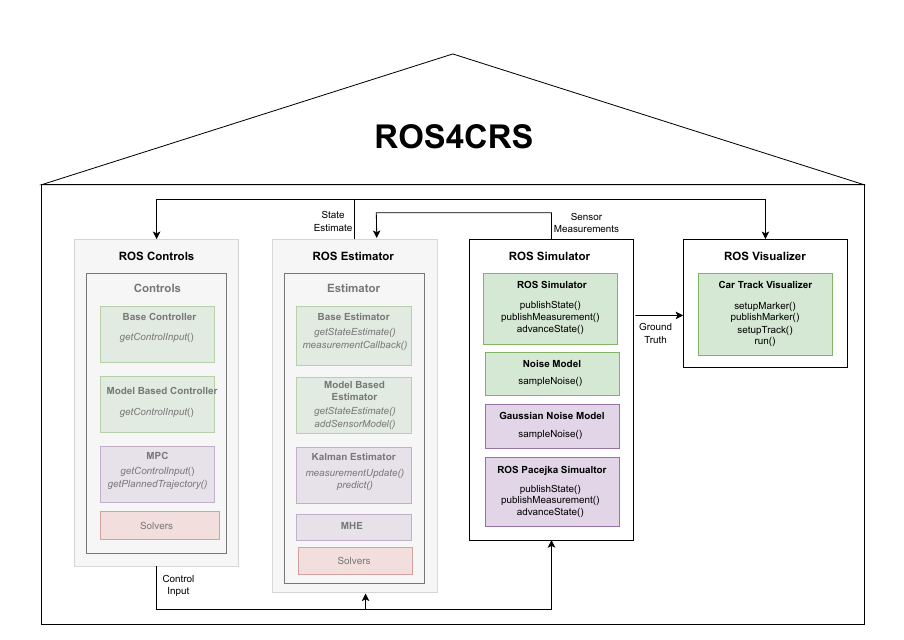}
    \caption{Overview of the ROS4CRS framework. The grey blocks demonstrate the ROS wrappers around the controller and estimator module. In addition, ROS4CRS provides a simulator as well as a visualizer. The arrows demonstrate the published information.}
    \label{fig:ros4crs_overview}
    \vspace{-15px}
\end{figure*}

\emph{ROS4CRS} consists of ROS wrappers for the controller and estimator module, as well as implementations for the simulator and visualizer. \cref{fig:ros4crs_overview} provides an overview, where the grey boxes demonstrate the ROS wrappers of the controller and estimator. The simulator provides (noisy) sensor measurements, while the visualizer provides markers that are visualized in RViz. The black arrows demonstrate the published information.

The split of the framework into \emph{CRS} and \emph{ROS4CRS} allows users to employ a different communication framework, such as migrating from ROS 1 to ROS 2 or any other framework, without the need to change the control pipeline implementation.

In addition, we provide \emph{tools} such as the optimization-based system identification package (\texttt{opt\_sys\_id}), which are mostly standalone and written in Python. The system identification package includes conversion tools from ROS bags for convenience.

\subsubsection{Hardware}

A custom-developed PCB based on the Espressif ESP32-WROOM-32D powers the Chronos small-scale car-like robots. The module offers a dual-core architecture with Wi-Fi and BLE connectivity. Chronos is designed for battery-powered applications, typically from rechargeable AAA batteries. To ensure a constant voltage supply to the drive and steer motors, a buck-boost circuit maintains a constant voltage over the entire state of charge of the batteries, enabling stable operation.

The steering and throttle DC motors are driven by two DRV8231A motor drivers connected to the MCU. On the sensor side, a BMI088 IMU provides high-quality inertial measurements, and an auxiliary STM32 MCU interfaces to four TMAG5123 Hall effect switches, which provide precisely timed interrupts when a wheel encoder magnet mounted in the wheel rim passes the switch. For steering position control, a rotary potentiometer mounted on the shaft provides an analog feedback voltage, sampled from an ADC on the STM32 chip. Total system current and battery voltage monitoring enable real-time power monitoring via the system telemetry.

The Lighthouse positioning deck is mounted on a header on the PCB, and attached to the ESP32's UART bus. The FPGA on the positioning deck decodes the bitstream from the base stations and provides up to 100 data frames per second to the ESP32.

\subsubsection{Firmware}

The firmware of the Chronos small-scale race car is written in C and C++, based on the ESP-IDF development framework. It makes use of the dual-core architecture by assigning its FreeRTOS task-based threads to core 0 (Wi-Fi and communication) and core 1 (embedded controllers and data acquisition). The telemetry rate of all sensor data matches the control loop rate at \SI{250}{\hertz}, while the commands received from the communication link may also be slower, depending on the update rate of the high-level controller in \emph{CRS}.

The low-level controllers are user-configurable: amongst other parameters, controller gains and thresholds are stored in non-volatile storage in the chip, and may be modified by enabling the car's configuration mode. In this state, the Chronos car will act as a wireless access point, where the parameters are modifiable through a web interface.

\subsubsection{Communication}

We leverage the built-in Wi-Fi module of the ESP32 chip and communicate via Wi-Fi, using a lightweight TCP/IP networking stack. All telemetry and control commands are sent via UDP, serialized using Protocol Buffers. An external Wi-Fi router connects the operating car(s) to the host computer.

